\pdfoutput=1

\documentclass[11pt]{article}

\usepackage[]{acl}

\usepackage{times}
\usepackage{latexsym}

\usepackage[T1]{fontenc}

\usepackage[utf8]{inputenc}

\usepackage{microtype}

\usepackage{amsthm}
\usepackage{amsmath}
\usepackage{amssymb}
\usepackage{caption}
\usepackage{subcaption}
\usepackage{enumerate}
\usepackage{multirow}
\usepackage{graphicx} 
\usepackage{url}
\usepackage{booktabs}
\usepackage{color}
\usepackage{algorithm}
\usepackage{algorithmic}
\usepackage{cleveref}
\crefname{section}{§}{§§}
\Crefname{section}{§}{§§}
\definecolor{newgreen}{RGB}{2,180,87} 

\title{Locate Then Ask: Interpretable Stepwise Reasoning for \\
Multi-hop Question Answering}

\author{Siyuan Wang\textsuperscript{\rm 1}, Zhongyu Wei\textsuperscript{\rm 1,2}\thanks{~~Corresponding author}, \bf Zhihao Fan\textsuperscript{\rm 1}, Qi Zhang\textsuperscript{\rm 3}, Xuanjing Huang\textsuperscript{\rm 3} \\
\textsuperscript{\rm 1}School of Data Science, Fudan University, China \\
\textsuperscript{\rm 2}Research Institute of Intelligent and Complex Systems, Fudan University, China \\
\textsuperscript{\rm 3}School of Computer Science, Fudan University, China \\
\{wangsy18,zywei,fanzh18,qz,xjhuang\}@fudan.edu.cn \\
}

\begin{document}
\maketitle
\begin{abstract}
Multi-hop reasoning requires aggregating multiple documents to answer a complex question. Existing methods usually decompose the multi-hop question into simpler single-hop questions to solve the problem for illustrating the explainable reasoning process. However, they ignore grounding on the supporting facts of each reasoning step, which tends to generate inaccurate decompositions.
In this paper, we propose an interpretable stepwise reasoning framework to incorporate both single-hop supporting sentence identification and single-hop question generation at each intermediate step, and utilize the inference of the current hop for the next until reasoning out the final result.
We employ a unified reader model for both intermediate hop reasoning and final hop inference and adopt joint optimization for more accurate and robust multi-hop reasoning. 
We conduct experiments on two benchmark datasets HotpotQA and 2WikiMultiHopQA. The results show that our method can effectively boost performance and also yields a better interpretable reasoning process without decomposition supervision. \footnote{Codes are publicly available at \url{https://github.com/WangsyGit/StepwiseQA}.} 
\end{abstract}

\section{Introduction}
Recent years have witnessed an emerging trend in the task of multi-hop question answering. It requires the model to aggregate multiple pieces of documents (i.e., context) and perform multi-hop reasoning to infer the answer~\cite{talmor2018web,khashabi2018looking}.
Several datasets have been introduced as benchmarks, such as HotpotQA~\cite{yang2018hotpotqa}, 2WikiMultiHopQA ~\cite{ho-etal-2020-constructing} and WikiHop~\cite{welbl2018constructing}, and the first two provide supporting facts supervision to encourage models to further explain what supporting sentences lead to the prediction.

\begin{figure}[!th]
\centering
\includegraphics[width=1.0\columnwidth]{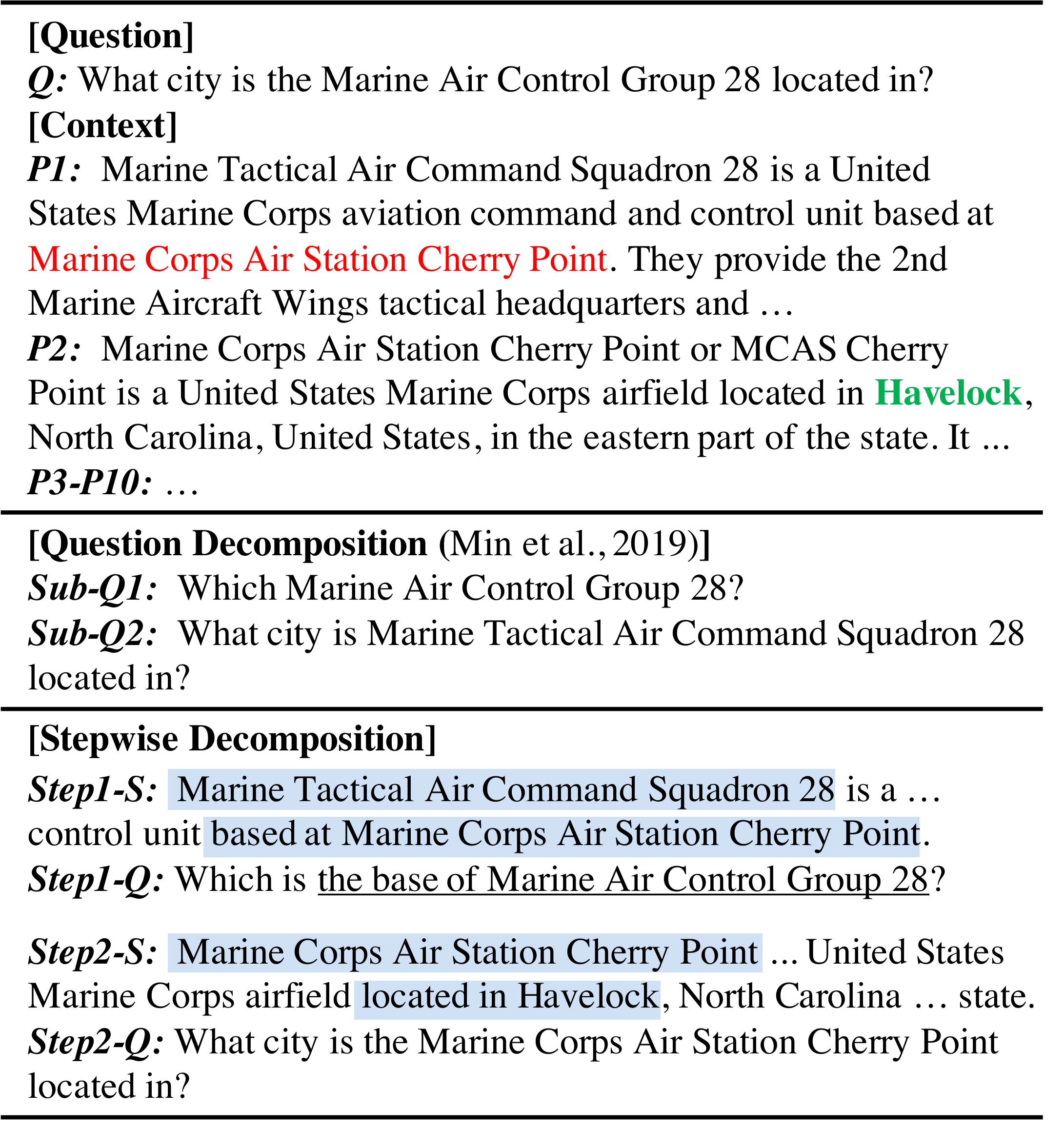}
\caption{\label{figure1} A multi-hop reasoning example from HotpotQA. To solve the problem, DecompRC \cite{min2019multi} generates improper decomposition of questions and predicts a \textcolor{red}{wrong answer} while our expected stepwise decomposition includes both single-hop supporting sentences and sub-questions of each step to reason out the \textcolor{newgreen}{correct answer}.
The underlined phrase is the fact uncovered by machine-generated decomposition while the shaded contexts support the corresponding single-hop question generation. 
}
\end{figure}

The first generation of models for multi-hop question answering utilizes a one-step reader~\cite{qiu2019dynamically,fang2019hierarchical,shao2020graph,tu2020select,beltagy2020longformer} to capture the interaction between the question and relevant contexts for the prediction of the answer as well as the supporting sentences. In order to model the explainable multi-step reasoning process, researchers explore to decompose the multi-hop question into easier single-hop questions and solve sub-questions to reach the answer~\cite{talmor2018web,wolfson2020break}. 

\begin{figure*}[!th]
\centering
\includegraphics[width=2.04\columnwidth]{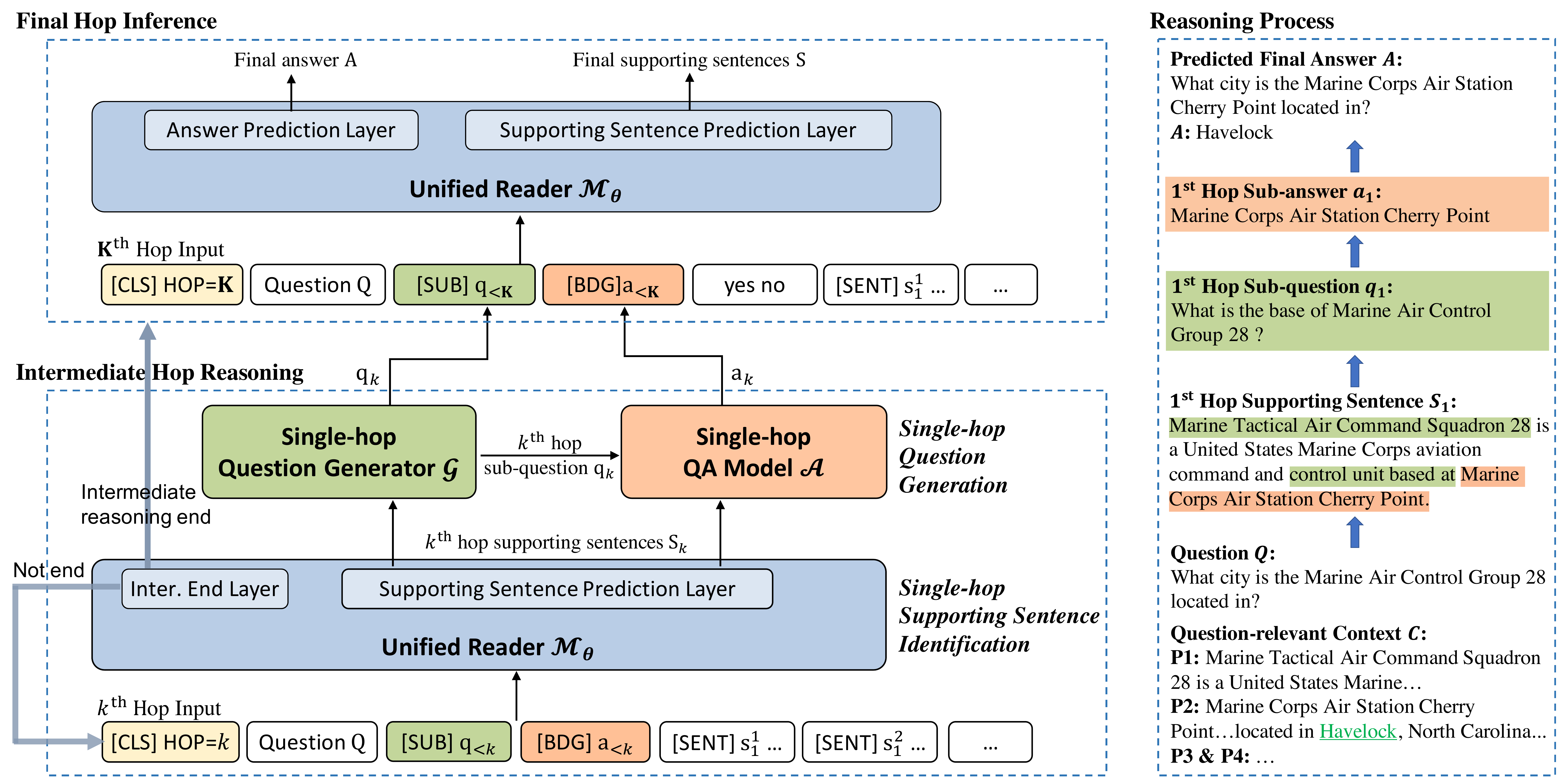}
\caption{\label{figure_framework} The overall architecture of stepwise reasoning framework with an interpretable reasoning process.
}
\end{figure*}
Question decomposition based approaches achieve promising prediction performance and are able to demonstrate the reasoning process to some extent. However, the single-hop questions are generated solely based on the original question without considering the supporting facts each step involves~\cite{min2019multi,perez2020unsupervised,khot2020text}. This usually leads to wrong-guided decomposition and inaccurate explanations. 
An example is shown in Figure~\ref{figure1} including a question, two relevant contextual paragraphs, two sub-questions generated by \citet{min2019multi} and two expected steps of reasoning with supporting sentences and sub-questions. 
The first single-hop question generated by an existing model (\emph{Sub-Q1}) fails to query ``the base of Marine Air Control Group 28'' which is beyond the scope of the original multi-hop question and such an improperly reasoned hop also leads to the failure of final answer prediction. We argue that a complete step of reasoning should consist of intermediate supporting sentence identification and sub-question generation to reduce the inference error in the procedure.

In this paper, we propose a stepwise reasoning framework for multi-hop question answering. It performs both single-hop supporting sentence identification and single-hop question generation in each step, and reasons from one intermediate hop to the next until the final hop inference. 
Specifically, we perform an intermediate hop reasoning that locates the single-hop supporting sentences and constructs the sub-question based on the original question and the corresponding supporting facts in each step.
We utilize an off-the-shelf single-hop question generator to eliminate the need for human annotations and avoid the risk of noisy labels posed by constructed pseudo-supervision.
In the final hop, we simultaneously predict the answer and the supporting sentences of the multi-hop question according to the preceding hops. We employ a unified reader model for both intermediate single-hop supporting sentence identification and final hop inference and jointly learn them so that a midway error may be corrected by subsequent hops to mitigate cumulative failures. 
We further adopt two measures to reduce the train-test discrepancy of single-hop supporting sentences and sub-questions to mitigate exposure bias for better generalization.

Experiments are conducted on two benchmark datasets involving different hops of reasoning, HotpotQA \cite{yang2018hotpotqa} and 2WikiMultiHopQA~\cite{ho-etal-2020-constructing}. The results indicate that our stepwise reasoning framework achieves significant improvements and shows general effectiveness across different reasoning types. 
Further analysis and qualitative cases also demonstrate that our method generates high-quality single-hop questions for interpretable multi-hop reasoning.

\section{Methodology}
Given a multi-hop question $Q$ and a context including multiple paragraphs, we aim to read the question-relevant context $C$ to predict the final answer $A$ and explain it with the supporting sentences $S$.
As illustrated in Figure~\ref{figure_framework}, we present a stepwise reasoning framework to iteratively identify the single-hop supporting sentences and generate the single-hop question for the following reasoning, 
which consists of three components as below. 
It first filters out the unrelated paragraphs to extract the question-relevant context $C$ (\cref{sec:paragraph_selection}).
Then it identifies the supporting sentences of each intermediate hop from the relevant context to ask and answer the corresponding single-hop question, and passes the auto-generated messages to the next hop (\cref{sec:intermediate_step}). After intermediate hop reasoning ends, the last module predicts the final answer and the supporting sentences of the multi-hop question according to the preceding inference (\cref{sec:final_step}). We jointly train a unified reader model for all reasoning hops (\cref{sec:joint_opt}) and adopt two measures to mitigate the train-test discrepancy for better inference (\cref{sec:exposure_bias}).

\subsection{Context Filter}
\label{sec:paragraph_selection}
In order to reduce the distraction in the context for downstream multi-hop reasoning process, we select the most relevant paragraphs as the question-relevant context $C$.
We first train a paragraph selection model which takes the question and the concatenation of all candidate paragraphs as the input and predicts the probability scores that each paragraph is relevant to answer the question. 

As HotpotQA mainly consists of 2-hop questions which involve two relevant paragraphs, we follow the 2-hop selection strategy in~\cite{fang2019hierarchical}.
For the first hop, it selects the paragraph with the highest scores among the paragraphs containing the same phrases as the question. Then the second-hop paragraph is extracted by hyperlinks from the first selected one. Two other paragraphs with the next highest scores are also selected to constitute the question-relevant context. 
For 2WikiMultiHopQA dataset with 2$\sim$4 hop questions, we select five paragraphs with the highest scores and filter the other paragraphs from the context.

\subsection{Intermediate Hop Reasoning}
\label{sec:intermediate_step}
Based upon the filtered relevant context, we perform the multi-hop reasoning step-by-step. We adopt a unified reader model $\mathcal{M}_{\theta}$ to iteratively identify the single-hop supporting sentences focused at each intermediate hop, and decide whether to end the intermediate hop reasoning indicating that the cumulative intermediate information is ready for final hop inference. Then depending on the predicted supporting sentences at each hop, the corresponding single-hop question can be generated and answered, which will be passed to the reader for next hop reasoning. 
This iterative process is repeated until the intermediate reasoning is ended, or 
up to $K-1$ intermediate hops.

\paragraph{Single-hop Supporting Sentence Identification}
The reader attempts to find the supporting sentences at each hop $k \in \{1,...,K-1\}$ given the concatenation of the original question $Q$, the sub question-answer pairs $\{(q_1, a_1) \ ... \ (q_{k-1}, a_{k-1})\}$ of previous hops, and the relevant context $C$ as the input.
Specifically, the concatenated sequence is formulated as 
\texttt{[CLS] HOP=k [SEP] Q [SUB] $q_1$ [BDG] $a_1$ ... [SUB] $q_{k-1}$ [BDG] $a_{k-1}$ [SEP]  [SENT] $\rm s^1_1$ [SENT] $\rm s^1_2$ ...[SEP] [SENT] $\rm s^2_1$ ...[SEP]} where \texttt{HOP=k} indicates the current hop number. Special tokens \texttt{[SUB]}, \texttt{[BDG]}, \texttt{[SENT]} respectively represent the single-hop sub-question, sub-answer and each sentence, and $s^i_j$ is the $j$-th sentence of the $i$-th paragraph in the relevant context $C$. 

On top of the representations of each sentence token \texttt{[SENT]}, we build a binary classifier to predict the probability $p^{(k)}_{i,j}$ that each sentence $s^i_j$ is a supporting fact of the current hop. The corresponding binary cross entropy loss $\mathcal{L}^{(k)}_{sf}$ is calculated as Eq.~\ref{eq_1}. $y^{(k)}_{i,j}$ is the label whether the sentence $s^i_j$ is a supporting fact of the hop $k$, and $N_s$ is the total number of sentences in $C$.
\begin{align}
\label{eq_1}
    \mathcal{L}^{(k)}_{sf} = \frac{1}{N_s}\sum_i &\sum_j -y^{(k)}_{i,j} \mathop{\log}(p^{(k)}_{i,j}) \nonumber \\
    &-(1-y^{(k)}_{i,j})\mathop{\log}(1-p^{(k)}_{i,j})
\end{align}
Then the \texttt{[CLS]} representation is also fed into a binary classifier to compute the probability $p^{(k)}_{end}$ that the intermediate reasoning should be ended at hop $k$ and go on to final hop inference, and the cross entropy loss is as Eq.~\ref{eq_2}. $y^{(k)}_{end}$ is the label whether to end the intermediate hop reasoning at current hop $k$.
\begin{align}
\label{eq_2}
    \mathcal{L}^{(k)}_{end} = -y^{(k)}_{end} &\mathop{\log}(p^{(k)}_{end}) \nonumber \\
    &-(1-y^{(k)}_{end})\mathop{\log}(1-p^{(k)}_{end})
\end{align}

\paragraph{Single-hop Question Generation}
After identifying the supporting sentences $S_k$ of hop $k$, we generate the corresponding single-hop question $q_k$ to investigate what the current hop asks about.
In this work, we do not use annotated or pseudo supervision to train a question decomposition model. Instead, we take inspiration from \cite{pan2020unsupervised} and adopt a pre-trained simple question generator $\mathcal{G}_s$ to directly output the desired single-hop question, which is trained beforehand on top of an off-the-shelf simple question corpus.

To encourage the single-hop question more grounded on the contextual facts, we generate them based on both the identified single-hop supporting sentences $S_k$ and the multi-hop question $Q$. The latter serves as a guidance for the generation towards the original reasoning goal. Specifically, we extract the intersectional tokens between $S_k$ and $Q$ as the prompt and append it to the supporting sentences $S_k$ as the input for single-hop question generation, which is organized as \texttt{[CLS] ($\rm Q \ \cap \ S_k$) [SEP] $\rm S_k$ [SEP]}.
Correspondingly, during the single-hop question generator pre-training, we also take a single sentence as the context and utilize the tokens existing within both the target question and the context to prompt simple question generation. 

Queried with the generated single-hop question $q_k$, we immediately resolve it to ease the whole multi-hop question. We also leverage the aforementioned simple question dataset to pre-train a single-hop QA model $\mathcal{A}$ to make it more consistent with the single-hop question generator. Then according to the single-hop supporting sentences $S_k$ and question $q_k$ at hop $k$, we utilize the pre-trained QA model $\mathcal{A}$ to output the single-hop answer $a_k$, which together with the single-hop question $q_k$ will be passed to the next hop for subsequent reasoning.

\subsection{Final Hop Inference}
\label{sec:final_step}
After the end of intermediate hop reasoning, we can utilize the single-hop questions and answers $\{(q_1, a_1) \ ... \ (q_{K-1}, a_{K-1})\}$ of all previous hops to build a bridge for inferring the final hop $K$. We employ the same unified reader model $\mathcal{M}_{\theta}$ during intermediate hop reasoning to predict the final answer $A$ of the multi-hop question $Q$ and simultaneously provide the overall explanatory supporting sentences $S$.
The input sequence fed into the reader for the final hop is similar to intermediate hops, except that we insert two additional tokens \texttt{yes} and \texttt{no} before the relevant context $C$ for answer prediction. As there are three answer types (yes/no/span), we integrate the answer type classification into the answer span prediction by appending \texttt{yes} and \texttt{no} as two candidate spans. We reformulate the current input as
\texttt{[CLS] HOP=K [SEP] Q [SUB] $q_1$ [BDG] $a_1$ ... [SUB] $q_{K-1}$ [BDG] $a_{K-1}$ [SEP] \textbf{yes no} [SEP] [SENT] $\rm s^1_1$ ...[SEP] [SENT] $\rm s^2_1$ ...[SEP]}. 

To accomplish the final hop inference, we first utilize the same binary classifier to identify whether each sentence is a supporting fact of the whole multi-hop question, and compute a final supporting sentence identification loss $\mathcal{L}^{K}_{sf}$.
Then for the final answer span prediction, we 
attach a linear layer with a softmax function to all context representations to obtain the probability of each token $t_n$ being the start $p^{s}_n$ or end position $p^{e}_n$ of the answer span.
The cross entropy loss is calculated as following formula, where token $t_{x}$
and $t_{y}$ are respectively the labels of start or end positions.
\begin{align}
    \mathcal{L}_{span} = - \mathop{\log}(p^{s}_{x}) -  \mathop{\log}(p^{e}_{y})
\end{align}

\subsection{Optimization \& Inference}
\label{sec:joint_opt}
In order to optimize our framework, we can first set up a maximum number of required reasoning hops $K$. Then our stepwise reasoning framework essentially comprises $K-1$ iterative intermediate hop reasoning layers and a final hop inference layer. As there is no previous single-hop question-answer pair before the first hop reasoning, we omit them within the initial input which will be fed into the first intermediate hop reasoning layer. 
We jointly train our stepwise reasoning framework for all intermediate hops and the final hop in order that an intermediate mistake can be corrected by subsequent hops to mitigate cascading failures.
All losses are combined in a weighted manner:
\begin{align}
    \mathcal{L} = &\lambda_1 \mathcal{L}^{int}_{sf}+\lambda_2 \mathcal{L}^{int}_{end}+\lambda_3 \mathcal{L}^{K}_{sf}+\mathcal{L}_{span} \\
    \mathcal{L}^{int}_{sf} &= \frac{\mathcal{L}^{(1)}_{sf}+\sum_{k=2}^{K-1} (1-y^{(k-1)}_{end})\mathcal{L}^{(k)}_{sf}}
    {1+\sum_{k=2}^{K-1} (1-y^{(k-1)}_{end})} \\
    \mathcal{L}^{int}_{end} &= \frac{1}{k_{e}} \sum_{k=1}^{k_{e}} \mathcal{L}^{(k)}_{end}, \text{where} \ y^{(k_e)}_{end}=1
\end{align}
where $\lambda_1$, $\lambda_2$ and $\lambda_3$ are weighted hyper-parameters. $\mathcal{L}^{int}_{sf}$ is the average supporting sentence identification loss of all actual intermediate hops that are not ended. $\mathcal{L}^{int}_{end}$ is the average loss of intermediate reasoning end prediction.

During the inference period, we start from the first hop and dynamically
reason from one hop to the next until the final hop. We predict whether to end the intermediate reasoning at each intermediate hop. Once it is over, we utilize all generated sub-questions and sub-answers and move to conduct the final hop inference. If not end, we will pass them to the next intermediate hop and repeat the process until the intermediate hop reasoning is ended, or until reaching $K-1$ intermediate hops.


\subsection{Exposure Bias Mitigation} 
\label{sec:exposure_bias}
In light of the design of our stepwise reasoning framework, there may arise the exposure bias problem between optimization and inference. Given the ground-truth supporting sentence supervision for each intermediate hop $S_k^t$, we can generate the target single-hop question and answer for the follow-up reasoning at training time. However, at test time we can only conduct single-hop question generation based on the predicted single-hop supporting sentences $S_k$ which may deviate from the oracle ones $S_k^t$. To address this, we propose two measures to respectively reduce the discrepancy between train-test single-hop supporting sentences and train-test single-hop questions. 

Firstly, we train a separate reader model only for the intermediate single-hop supporting sentence identification, and adopt it to re-predict the single-hop supporting sentences of training data with occasionally injected mistakes for optimizing the whole framework. Thereby we can regulate bias between training and test supporting sentences of intermediate hops. Besides, we also augment the training data for the single-hop question generation $\mathcal{G}_s$ by taking the re-predicted training single-hop supporting sentences $S_k$ as input, and the generated sub-questions based on the ground-truth supporting sentences $S_k^t$ as the target. Then the generator is trained to recover from the non-gold single-hop supporting sentences to approximate the oracle ones and reduce the deviation between train-test intermediate single-hop questions.
With these two strategies, we can jointly optimize our stepwise reasoning framework for better generalization to non-golden test cases.

\section{Experiments}
\subsection{Experimental Dataset}
\label{implemental}
We take two datasets HotpotQA~\cite{yang2018hotpotqa} and 2WikiMultiHopQA~\cite{ho-etal-2020-constructing} that involve different reasoning hops as a testbed to study textual multi-hop reasoning. They both require answering the question as well as predicting the supporting facts to explain the reasoning. HotpotQA includes
both distractor setting and fullwiki setting, and we focus on the former with limited candidate paragraphs to fully test the multi-hop reasoning ability while putting aside the information retrieval part. Although 2WikiMultiHopQA provides annotated evidence for interpreting the reasoning path, we leave them out of account to illustrate the effectiveness and interpretability of our framework without explanation supervision.

HotpotQA and 2WikiMultiHopQA respectively consist of $90,447/7,405/7,405$ and $167,454/12,576/12,576$ samples in training, development and test sets,
and each instance is provided with 10 paragraphs. HotpotQA comprises 2-hop questions that only two paragraphs contain necessary supporting sentences, while 2WikiMultiHopQA contains 2$\sim$4 hop questions and the supporting facts reside in two to four paragraphs.
Besides, to train the single-hop question generator and QA model, we use SQuAD~\cite{rajpurkar2016squad} as the simple question corpus.

\begin{table*}[ht]
\begin{center}
\resizebox{0.72\textwidth}{!}{
\begin{tabular}{lcccccc}
\toprule
\multirow{2}{*}[-0.7ex]{\centering Model}
&\multicolumn{2}{c}{Answer} &\multicolumn{2}{c}{Sup Fact} &\multicolumn{2}{c}{Joint} \\
\cmidrule(lr){2-3}\cmidrule(lr){4-5}\cmidrule(lr){6-7}
 & EM & F1 & EM & F1 & EM & F1 \\
\midrule
\emph{DecompRC}~\cite{min2019multi} & 55.20 & 69.63 & - & - & - & - \\
\emph{ONUS}~\cite{perez2020unsupervised} & 66.33 & 79.34 & - & - & - & - \\
\midrule
\emph{TAP2}~\cite{glass2019span} & 64.99 & 78.59 & 55.47 & 85.57 & 39.77 & 69.12 \\
\emph{SAE-large}~\cite{tu2020select}& 66.92 & 79.62 & 61.53 & 86.86 & 45.36 & 71.45 \\
\emph{C2F Reader}~\cite{shao2020graph} & 67.98 & 81.24 & 60.81 & 87.63 & 44.67 & 72.73 \\
\emph{Longformer}~\cite{beltagy2020longformer} & 68.00 & 81.25 & 63.09 & 88.34 & 45.91 & 73.16 \\
\emph{ETC-large}~\cite{zaheer2020big} & 68.12 & 81.18 & \bf 63.25 & \bf 89.09 & 46.40 & 73.62 \\
\emph{FFReader-large}~\cite{alkhaldi2021flexibly} & 68.89 & 82.16 & 62.10 & 88.42 &  45.61 & 73.78 \\
\emph{HGN-large}~\cite{fang2019hierarchical} & 69.22 & 82.19 & 62.76 & 88.47 & 47.11 & 74.21 \\
\midrule
\emph{StepReasoner} & \bf 69.66 & \bf 82.42 & 62.99 & 87.85 & \bf 47.84 & \bf 74.27 \\
\bottomrule
\end{tabular}
}
\caption{\label{table_result_1} Experimental results of different models on the test set of HotpotQA distractor setting. 
}
\end{center}
\end{table*}

\subsection{Implementation Details}
\label{implementation}
We take ELECTRA-large~\cite{clark2020electra} as the backbone of our proposed framework and the single-hop QA model, and train a single-hop question generator using BART-large~\cite{lewis2019bart}. 
The weights to balance losses are chosen as $\lambda_1 = 10/5$ (for HotpotQA/2WikiMultiHopQA), $\lambda_2 = 2$ and $\lambda_3 = 5$. The maximum value of required reasoning hops is set as $K=2$ for HotpotQA and $K=4$ for 2WikiMultiHopQA. More training details are given in Appendix~\ref{training_details}.

\subsection{Overall Performance}
We compare our \textbf{step}wise \textbf{reasoner} (\emph{StepReasoner}) with previous published methods on HotpotQA and 2WikiMultiHopQA. 
Since there are few systems on the leaderboard of 2WikiMultiHopQA, we follow \cite{fu-etal-2021-decomposing-complex} and make a comparison with more models on both dev and test sets. 
The compared methods cover both question decomposition based models and one-step reading based models. 
\begin{table}[!th]
	\centering
    \begin{center}
    \resizebox{0.49\textwidth}{!}{
    \setlength{\tabcolsep}{1.8mm}{
    \begin{tabular}{l|cc|cc}
    \toprule
    \multirow{2}{*}[-0.7ex]{\centering Model}
    &\multicolumn{2}{c}{Answer} 
    &\multicolumn{2}{c}{Sup Fact} \\
    \cmidrule(lr){2-3}\cmidrule(lr){4-5} & EM & F1
    & EM & F1 \\
    \midrule
    \midrule 
    \multicolumn{5}{c}{Dev} \\
    \midrule
    \emph{DecompRC}~\cite{min2019multi} & 7.46 & 41.57 & 56.49 & 82.73 \\
    \emph{QFE}~\cite{nishida2019answering} & 37.56 & 43.21 & 21.13 & 59.20 \\
    \emph{CRERC}~\cite{fu-etal-2021-decomposing-complex} & 71.56 & 74.51 & \bf 86.00 & \bf 92.75 \\
    \emph{NA-Reviewer}~\cite{fu2022reviewer} & 76.88 & 82.30 & - & - \\
    \midrule
    \emph{DFGN}~\cite{qiu2019dynamically} & 30.87 & 38.49 & 17.06 & 57.79 \\
    \emph{ELECTRA-base} & 66.81 & 72.28 & 81.19 & 90.96 \\
    \emph{ELECTRA-large} & 79.22 & 83.51 & 83.08 & 92.01 \\
    \midrule
    \emph{StepReasoner}(ELECTRA-base) & 68.11 & 73.03 & 81.72 & 91.21 \\
    \emph{StepReasoner}(ELECTRA-large) & \bf 80.23 & \bf 84.26 & 83.41 & 92.01 \\
    \midrule
    \midrule
    \multicolumn{5}{c}{Test} \\
    \midrule
    \emph{HGN-revise}~\cite{fang2019hierarchical} & 71.20 & 75.69 & 69.35 & 89.07 \\
    \emph{CRERC}~\cite{fu-etal-2021-decomposing-complex} & 69.58 & 72.33 & 82.86 & 90.68 \\
    \emph{NA-Reviewer}~\cite{fu2022reviewer} & 76.73 & 81.91 & \bf 89.61 & \bf 94.31 \\
    \emph{StepReasoner} & \bf 80.88 & \bf 84.86 & 83.30 & 91.89 \\
    \bottomrule
    \end{tabular}
    }
    }
    \caption{\label{table_2wiki} Results on the dev and test sets of 2WikiMultiHopQA.}
    \end{center}
\end{table}

\noindent\textbf{Question decomposition based models}\ include \emph{DecompRC}~\cite{min2019multi}, \emph{ONUS}~\cite{perez2020unsupervised}, \emph{QFE}~\cite{nishida2019answering}, \emph{CRERC}~\cite{fu-etal-2021-decomposing-complex}) and \emph{NA-Reviewer}~\cite{fu2022reviewer}.

\noindent\textbf{One-step reading based models}\ consist of \emph{DFGN}~\cite{qiu2019dynamically}, \emph{ELECTRA}~\cite{clark2020electra}, \emph{TAP2}~\cite{glass2019span},
\emph{SAE-large}~\cite{tu2020select}, \emph{C2F Reader}~\cite{shao2020graph}, \emph{Longformer}~\cite{beltagy2020longformer}, \emph{ETC-large}~\cite{zaheer2020big}, \emph{FFReader-large}~\cite{alkhaldi2021flexibly}, \emph{HGN-large}~\cite{fang2019hierarchical}.

The results are shown in Table~\ref{table_result_1} and~\ref{table_2wiki}. We find that \emph{StepReasoner} outperforms all models in terms of both
answer prediction and joint evaluation and achieves comparable performance in supporting fact prediction, which demonstrates the effectiveness of our method. Specifically, it performs better than both question decomposition based and one-step reading based methods. The former improvement indicates that a unified reader to stepwise identify the single-hop supporting sentences for single-hop sub-question generation can enhance the accuracy of previous question decomposition methods. The latter verifies that the interpretability injected by stepwise reasoning can also improve the QA performance. Besides, \emph{Longformer}, \emph{ETC-large}, \emph{FFReader-large} and \emph{HGN-large} show a better supporting fact prediction performance on HotpotQA, especially in F1 score. This is because the first three models are designed for handling longer sequences and the last utilizes a complex hierarchical graph network which both can cover more candidate paragraphs for supporting sentence prediction. For 2WikiMultiHopQA, \emph{CRERC} and \emph{NA-Reviewer} achieve better supporting fact prediction because they both utilize external annotated evidence for training which confirms the effectiveness of our \emph{StepReasoner} without explanation supervision. We also evaluate the intermediate reasoning end prediction, and find that our \emph{StepReasoner} can exactly decide when to end the intermediate hop reasoning.

\subsection{Further Analysis}
\label{further_analysis}
\paragraph{Ablation Study} 
To dive into the sources of performance gain in our \emph{StepReasoner}, we conduct an ablation study on the HotpotQA development set, which is shown in Table~\ref{table_ablation}. Compared to the overall stepwise reasoning system \emph{StepReasoner}, a pipeline model without joint training shows a sharp performance degradation. It indicates that joint optimization of a unified reader model for all hops can improve the tolerance for intermediate faults and boost reasoning performance.
After removing any measures to mitigate exposure bias, the performance has also significantly dropped. It shows that both two measures to alleviate the train-test discrepancy of single-hop supporting sentences and single-hop questions confirm a better generalization to cases deviated from oracle.
\begin{table}[!th]
	\centering
    \begin{center}
    \resizebox{0.5\textwidth}{!}{
    \setlength{\tabcolsep}{1.8mm}{
    \begin{tabular}{l|cc|cc|cc}
    \toprule
    \multirow{2}{*}[-0.7ex]{\centering Model}
    &\multicolumn{2}{c}{Answer} 
    &\multicolumn{2}{c}{Sup Fact} 
    &\multicolumn{2}{c}{Joint} \\
    \cmidrule(lr){2-3}\cmidrule(lr){4-5} \cmidrule(lr){6-7} & EM & F1
    & EM & F1 & EM & F1 \\
    \midrule
    \emph{StepReasoner} & 70.11 & 83.03 & 64.27 & 88.10 & 48.55 & 74.85 \\
    \ \ \ \ \emph{w/o joint training}  & 69.30 & 82.44 & 63.35 & 87.89 & 47.25 & 74.16  \\
    \ \ \ \ \emph{w/o bias.supp}  & 69.66 & 82.64 & 63.10 & 87.74 & 47.39 & 74.20  \\
    \ \ \ \ \emph{w/o bias.ques}  & 69.76 & 82.93 & 63.46 & 88.01 & 47.49 & 74.57  \\
    \bottomrule
    \end{tabular}
    }
    }
    \caption{\label{table_ablation} Ablation study on HotpotQA dev set. \emph{w/o joint training} means a pipeline stepwise reasoning schema. \emph{bias.supp} and \emph{bias.ques} are two exposure \textbf{bias} mitigating measures to reduce the train-test discrepancy of single-hop \textbf{supp}orting sentences and \textbf{ques}tions.}
    \end{center}
\end{table}

\paragraph{Effectiveness on Various Backbone Models} To analyze the effectiveness of the \emph{StepReasoner} based on different backbones, we vary several pre-trained models of different scales including \emph{BERT-base-uncased}~\cite{liu2019roberta}, \emph{ELECTRA-large} and \emph{ALBERT-xxlarge-v2}~\cite{lan2019albert}. From Table~\ref{table_different_base}, we can see that our \emph{StepReasoner} variants consistently perform better than the corresponding baseline models and the previous state-of-the-art method (HGN) using \emph{ALBERT-large} as base model, especially in EM scores. It demonstrates that our method is robust to be effective based on various pre-trained models and it is the paradigm of our joint stepwise reasoning that contributes to more accurate multi-hop reasoning.

\begin{table}[!th]
	\centering
    \begin{center}
    \resizebox{0.5\textwidth}{!}{
    \setlength{\tabcolsep}{1.8mm}{
    \begin{tabular}{l|cc|cc|cc}
    \toprule
    \multirow{2}{*}[-0.7ex]{\centering Model}
    &\multicolumn{2}{c}{Answer} 
    &\multicolumn{2}{c}{Sup Fact} 
    &\multicolumn{2}{c}{Joint} \\
    \cmidrule(lr){2-3}\cmidrule(lr){4-5} \cmidrule(lr){6-7} & EM & F1
    & EM & F1 & EM & F1 \\
    \midrule    
    \emph{BERT} & 60.80& 74.76 & 57.16 & 85.05 & 38.67 & 65.89 \\
    \emph{StepReasoner}(BERT) & 60.52 & 74.81 & 59.00 & 85.38 & 40.31 & 66.50 \\
    \midrule
    \emph{ELECTRA} & 69.49 & 82.76 & 62.80 & 87.91 & 46.75 & 74.33 \\
    \emph{StepReasoner}(ELECTRA) & 70.11 & 83.03 & 64.27 & 88.10 & 48.55 & 74.85 \\
    \midrule
    \emph{ALBERT} & 70.14 & 83.58 & 62.78 & 88.43 & 46.60 & 75.30 \\
    \emph{HGN}(ALBERT) & 70.18 & 83.44 & 63.17 & 89.19 & 47.01 & 75.74 \\
    \emph{StepReasoner}(ALBERT) & 70.73 & 83.92 & 64.17 & 88.69 & 48.54 & 75.85 \\
    \bottomrule
    \end{tabular}
    }
    }
    \caption{\label{table_different_base} Analysis of \emph{StepReasoner} on different backbone models on HotpotQA dev set.}
    \end{center}
\end{table}

\paragraph{Analysis of Different Reasoning Types}
We detailedly investigate the performance of \emph{StepReasoner} on various reasoning types of HotpotQA compared to the baseline model. Following~\citet{min2019multi}, HotpotQA integrates four types of multi-hop reasoning skills, including ``Bridge'', ``Implicit-Bridge'', ``Comparison'' and ``Intersection''. The first two both require identifying the bridge entity to complete the chain reasoning, but ``Implicit-Bridge'' resembles single-hop questions which implicitly query a multi-hop property of an entity, such as the question in Fig.~\ref{figure1}. ``Intersection'' questions ask to locate the answer entity that satisfies multiple properties. ``Comparison'' questions involve comparing two entities to find the answer.

\begin{table}[!h]
	\begin{center}
	\resizebox{0.44\textwidth}{!}{
	\begin{tabular}{lcccc}
	\toprule
	\multirow{2}{*}[-0.7ex]{Reasoning Type}
    &\multicolumn{2}{c}{\emph{ELECTRA}} 
    &\multicolumn{2}{c}{\emph{StepReasoner}} \\
    \cmidrule(lr){2-3}\cmidrule(lr){4-5} & EM & F1
    & EM & F1 \\
	\midrule  				
	Bridge (34\%) & 47.30 & 76.43 & 48.37 & 76.54 \\
	Implicit-Bridge (29\%) & 37.04 & 68.66 & 39.81 & 69.65 \\  
	Comparison (20\%) & 61.77 & 79.13 & 63.04 & 79.13 \\
	Intersection (17\%) & 42.97 & 73.89 & 46.23 & 75.07 \\
	\bottomrule
 	\end{tabular}
 	}
 	\caption{\label{table_type}Results of Joint EM and Joint F1 across different reasoning types. The numbers in parentheses are percentages of different types. 
 	}
	\end{center}
\end{table}
As shown in Table~\ref{table_type}, our system is generally effective on all reasoning types compared to the baseline model \emph{ELECTRA}, especially ``Implicit-Bridge'' and ``Intersection''. Because these questions are susceptible to shortcut solutions by directly identifying an entity satisfying one queried property from a single piece of evidence to reach the incorrect answer while ignoring the multi-hop reasoning involving other evidence.
This observation also verifies the effectiveness of our system to stepwise generate the single-hop question grounded on the intermediate single-hop supporting sentences for interpretable multi-hop reasoning.  

\paragraph{Comparison of Different Single-hop Question Generation Methods}
To manifest the effectiveness of our generated single-hop questions based on identified single-hop supporting sentences, we incorporate several various single-hop question generation approaches into our stepwise reasoning framework and compare the QA results. Table~\ref{table_decomposition} shows that our \emph{Supp-based} method performs best.
It reveals that our single-hop question generation is grounded on single-hop supporting sentences to generate more accurate and informative sub-questions, which are more effective than single-hop questions constructed by other strategies.
\begin{table}[!th]
	\centering
    \begin{center}
    \resizebox{0.48\textwidth}{!}{
    \setlength{\tabcolsep}{1.8mm}{
    \begin{tabular}{l|cc|cc|cc}
    \toprule
    \multirow{2}{*}[-0.7ex]{\centering Method}
    &\multicolumn{2}{c}{Answer} 
    &\multicolumn{2}{c}{Sup Fact} 
    &\multicolumn{2}{c}{Joint} \\
    \cmidrule(lr){2-3}\cmidrule(lr){4-5} \cmidrule(lr){6-7} & EM & F1
    & EM & F1 & EM & F1 \\
    \midrule    
    \emph{Span-based} & 68.60 & 81.66 & 62.31 & 87.41 & 46.26 & 73.20 \\
    \emph{USeq2Seq} & 69.29 & 82.22 & 63.11 & 88.00 & 46.96 & 74.08 \\
    \emph{Supp-based} & 70.11 & 83.03 & 64.27 & 88.10 & 48.55 & 74.85 \\
    \bottomrule
    \end{tabular}
    }
    }
    \caption{\label{table_decomposition} Comparison of various single-hop question generation methods. \emph{Span-based} represents the sub-question generation \underline{\bf based} on \underline{\bf span} prediction in \emph{DecompRC}~\cite{min2019multi} while \emph{USeq2Seq} is the \underline{\bf U}nsupervised \underline{\bf seq2seq} decomposition in \emph{ONUS}~\cite{perez2020unsupervised}. \emph{Supp-based} is our \underline{\bf supp}orting sentences \underline{\bf based} single-hop question generation.}
    \end{center}
\end{table}

\begin{figure}[!th]
\centering
\includegraphics[width=1\columnwidth]{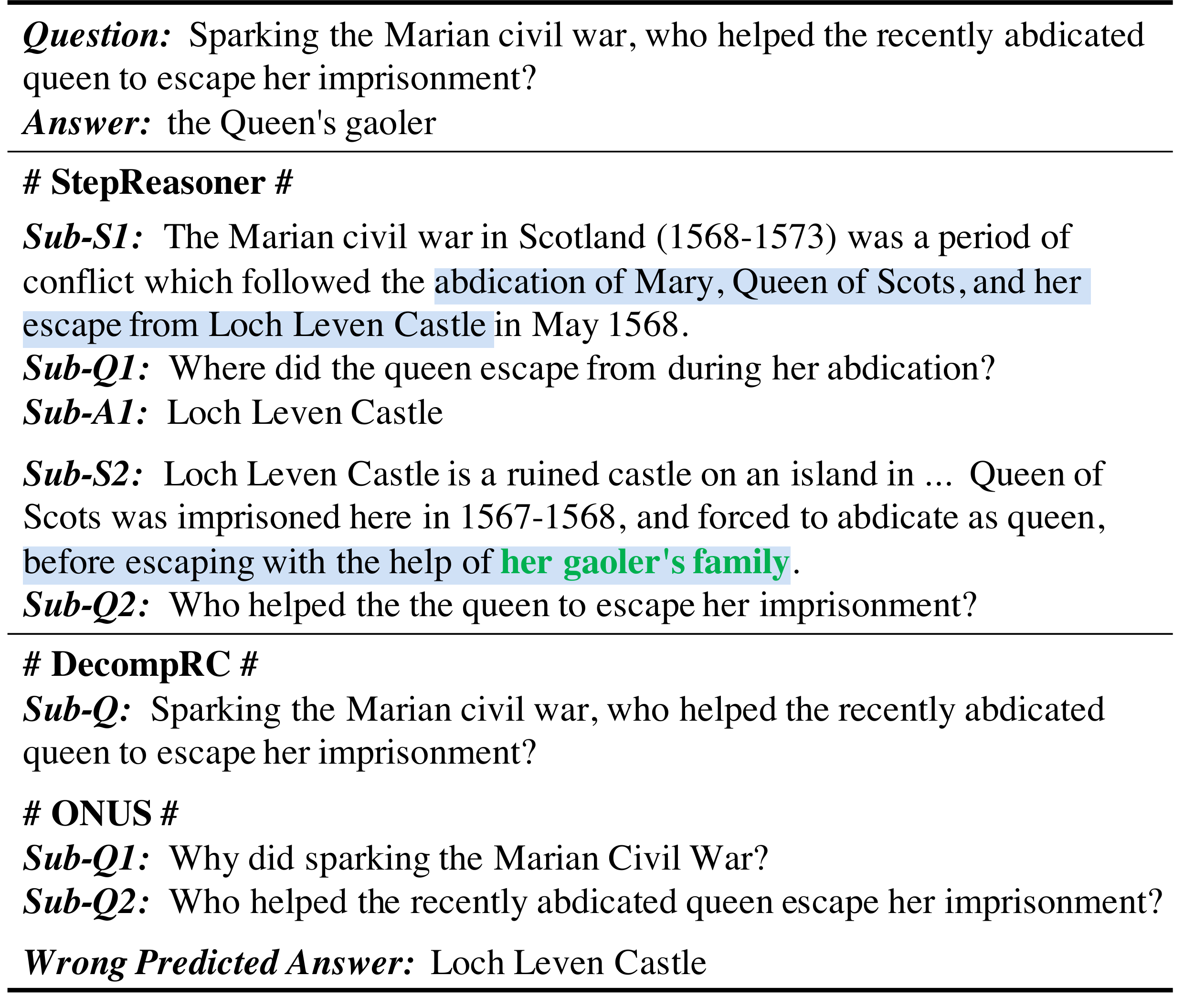}
\caption{\label{figure_case} A case of the reasoning process by our \emph{StepReasoner} compared to \emph{DecompRC} and \emph{ONUS}. The \textcolor{newgreen}{green} phrase denotes our predicted answer and the texts in shadow support the single-hop question generation. 
}
\end{figure}

\paragraph{Case Study}
An example of the ``Bridge'' type question is presented in Figure~\ref{figure_case} to show the interpretable reasoning process of \emph{StepReasoner} compared to other decomposition based methods. Our system successfully identifies the first-hop supporting sentences and generates the first-hop sub-question to query the escaping location. Then the first-hop sub-answer helps to identify the following supporting sentences
to finally predict the correct answer. The second-hop question is also generated for better illustration. By contrast, \emph{DecompRC} fails to decompose this complex question while \emph{ONUS} generates an improper first-hop question, and they both predict a wrong answer. More cases of the other reasoning types are in Appendix~\ref{more_cases}.

\section{Related Work}
Multi-hop question answering aims to aggregate multiple pieces of documents to model the multi-hop reasoning chain and predict the answer~\cite{khashabi2018looking,yang2018hotpotqa,welbl2018constructing,ho-etal-2020-constructing}.
Prior methods 
mainly focus on utilizing a single reader to model the interaction between the question and relevant context, and simultaneously or separately predict the supporting sentences and answer within one step. \citet{dhingra2018neural} and \citet{qiu2019dynamically} propose to construct graphs based on entity information from scattered paragraphs and utilize graph neural networks as the reader to reason out the answer. Then HDE-Graph~\cite{tu2019multi}, HGN~\cite{fang2019hierarchical} and SAE~\cite{tu2020select} enrich information in the entity graph by extending nodes of other granularity to build a hierarchical graph and improve the interaction between the question and context. Some other methods~\cite{glass2019span,beltagy2020longformer,zaheer2020big,alkhaldi2021flexibly} adopt pre-trained models as the powerful reader for multi-hop reasoning and achieve promising results.
However, these one-step reader methods directly encode the relevant context and question for answer prediction while neglecting to illustrate the explicit reasoning process.

To circumvent the interpretability limitation, another stream of research proposes to solve the multi-hop reasoning by multi-step question decomposition. \citet{nishida2019answering} and \citet{jiang2019self} recurrently update the sub-query at each step to break down the problem but these sub-queries are learned in latent representations and not sufficiently explainable. Instead, some works explore to explicitly decompose the complex question into single-hop questions which assumes access to decomposition supervision~\cite{min2019multi,wolfson2020break}. To skip this reliance, \cite{perez2020unsupervised} and \cite{khot2020text} attempt to construct pseudo-decomposition from other simple question corpora which pose a new challenge of label noises. Besides, they only take as input the original question to generate single-hop questions without grounding on the supporting facts at each hop. In contrast, we design a stepwise reasoning framework to locate the single-hop supporting sentences at each step for generating more fact-grounded and informative single-hop sub-questions without any genuine or pseudo supervision, and integrate the sequential reasoning process into a unified multi-hop reader for more robust performance.

\section{Conclusion}
In this paper, we study the task of multi-hop question answering and propose to stepwise locate the single-hop supporting sentences and generate more fact-grounded single-hop questions for better interpretable multi-hop reasoning. We present a stepwise reasoning framework to incorporate both single-hop supporting sentence identification and the corresponding single-hop question generation for each intermediate step until inferring a final result. 
It employs a pre-trained simple question generator and takes the identified single-hop supporting sentences as base to generate the single-hop question, which obviates the necessity of constructed supervision and helps generate more fact-based single-hop questions. It utilizes a unified reader to jointly learn both intermediate hop reasoning and final hop inference for better fault tolerance.
Experimental results validate the general effectiveness and interpretability of our \emph{StepReasoner}.


\bibliography{anthology,coling}
\bibliographystyle{acl_natbib}

\appendix

\section{Training Details}
\label{training_details}
All these models are implemented using Huggingface~\cite{wolf2019huggingface}. 
For HotpotQA, we use a batch size of 48 and fine-tune for 10 epochs with the learning rate 3e-5. For 2WikiMultiHopQA, the batch size is set to 24, the number of training epochs is 5 and the learning rate is 5e-5.
The Adam is taken as the optimizer and we use a linear learning rate scheduler with $10\%$ warmup proportion. 
The proposed systems and other comparison models are trained on 4 NVIDIA Tesla V100 GPUs.

\section{Case Study of Different Reasoning Types}
\label{more_cases}
\begin{figure}[!th]
  \centering
  \begin{minipage}[h]{0.5\textwidth}
    \centering
    \includegraphics[width=1.0\columnwidth]{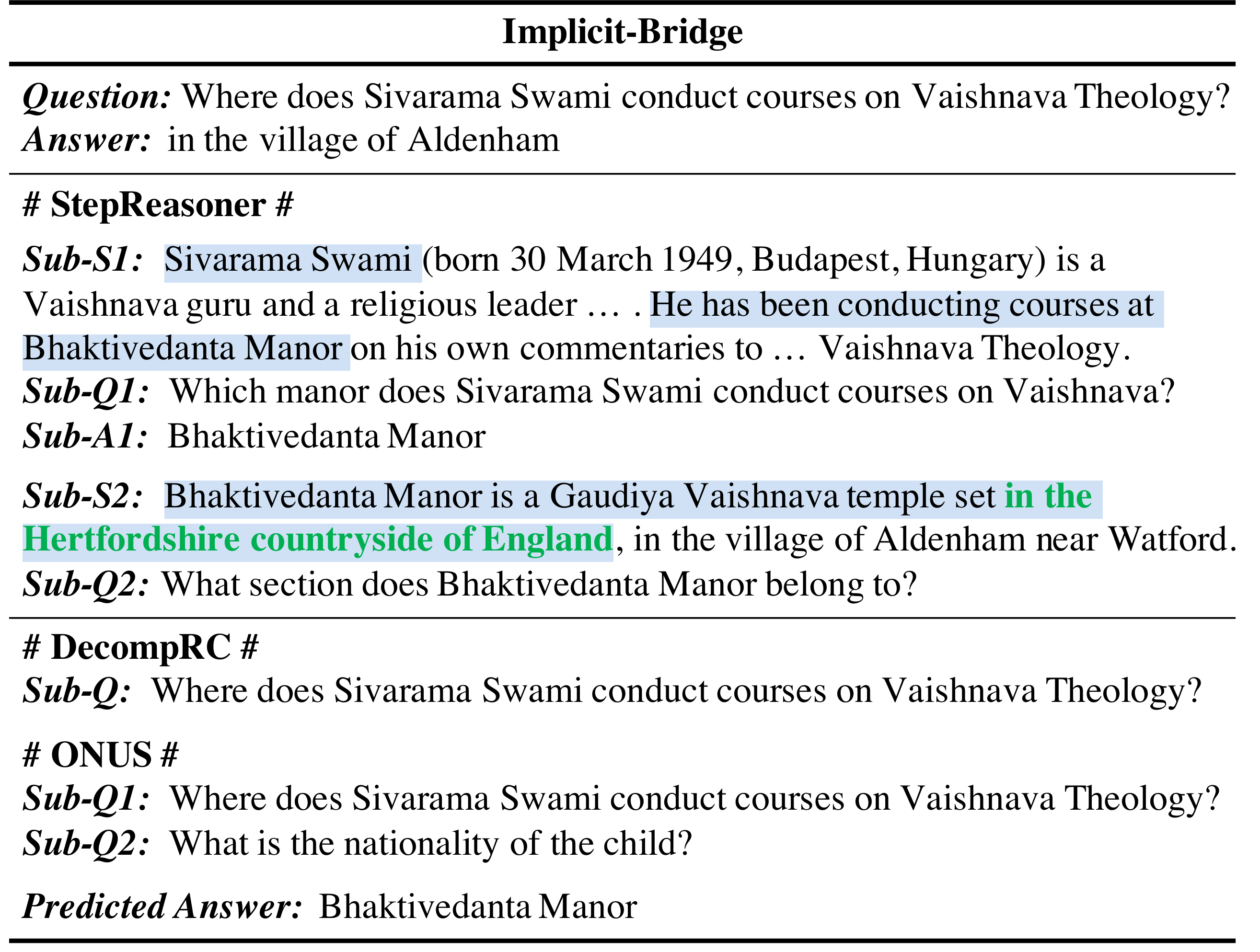}
    \subcaption{An example of ``Implicit-Bridge'' reasoning type.} 
    \label{figure1a}
  \end{minipage}

  \vspace{5mm}
  \begin{minipage}[h]{0.5\textwidth}
    \centering
    \includegraphics[width=1.0\columnwidth]{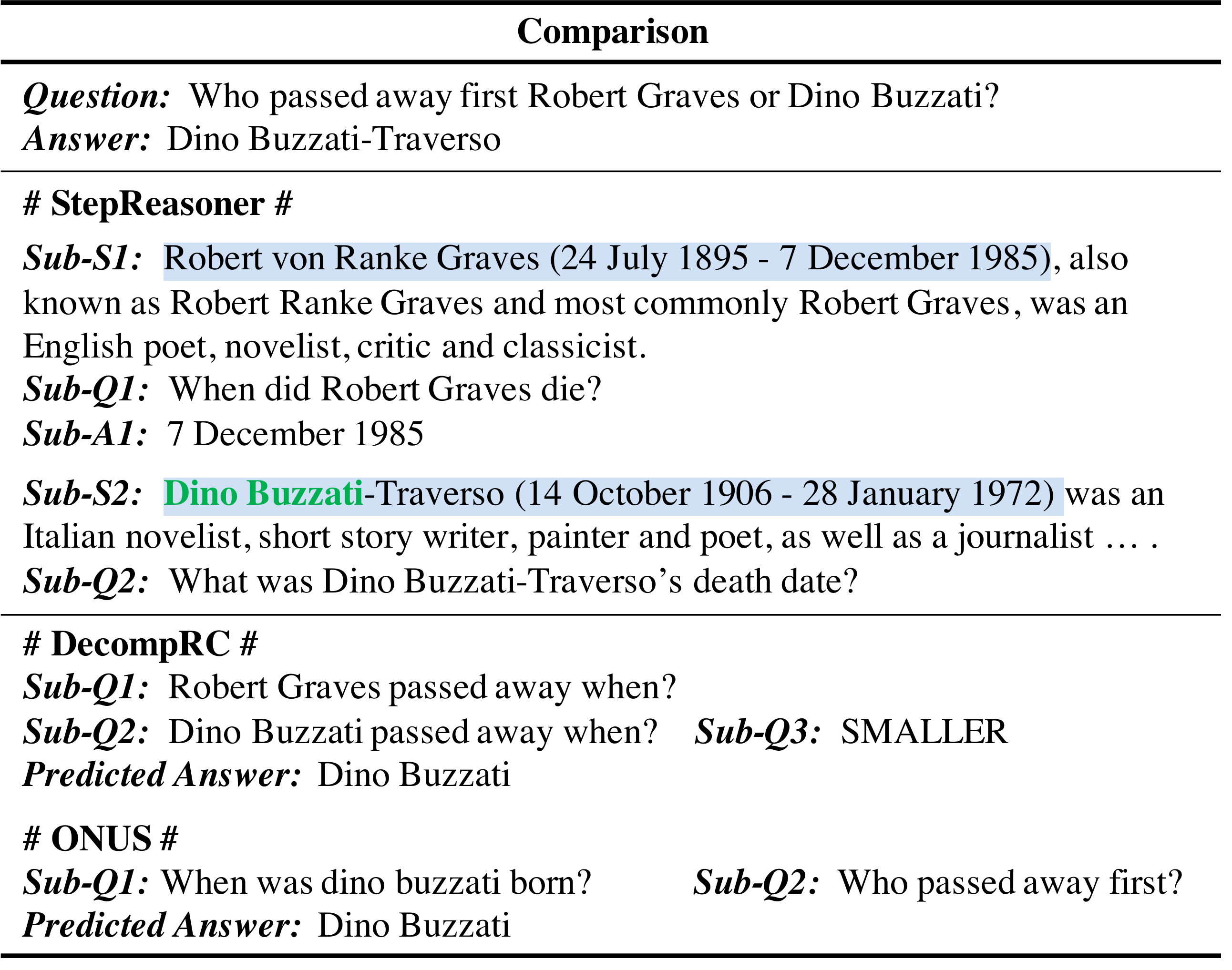} 
    \subcaption{An example of ``Comparison'' reasoning type.} 
    \label{figure1b}
  \end{minipage}
  
\vspace{5mm}
  \begin{minipage}[h]{0.5\textwidth}
    \centering
    \includegraphics[width=1.0\columnwidth]{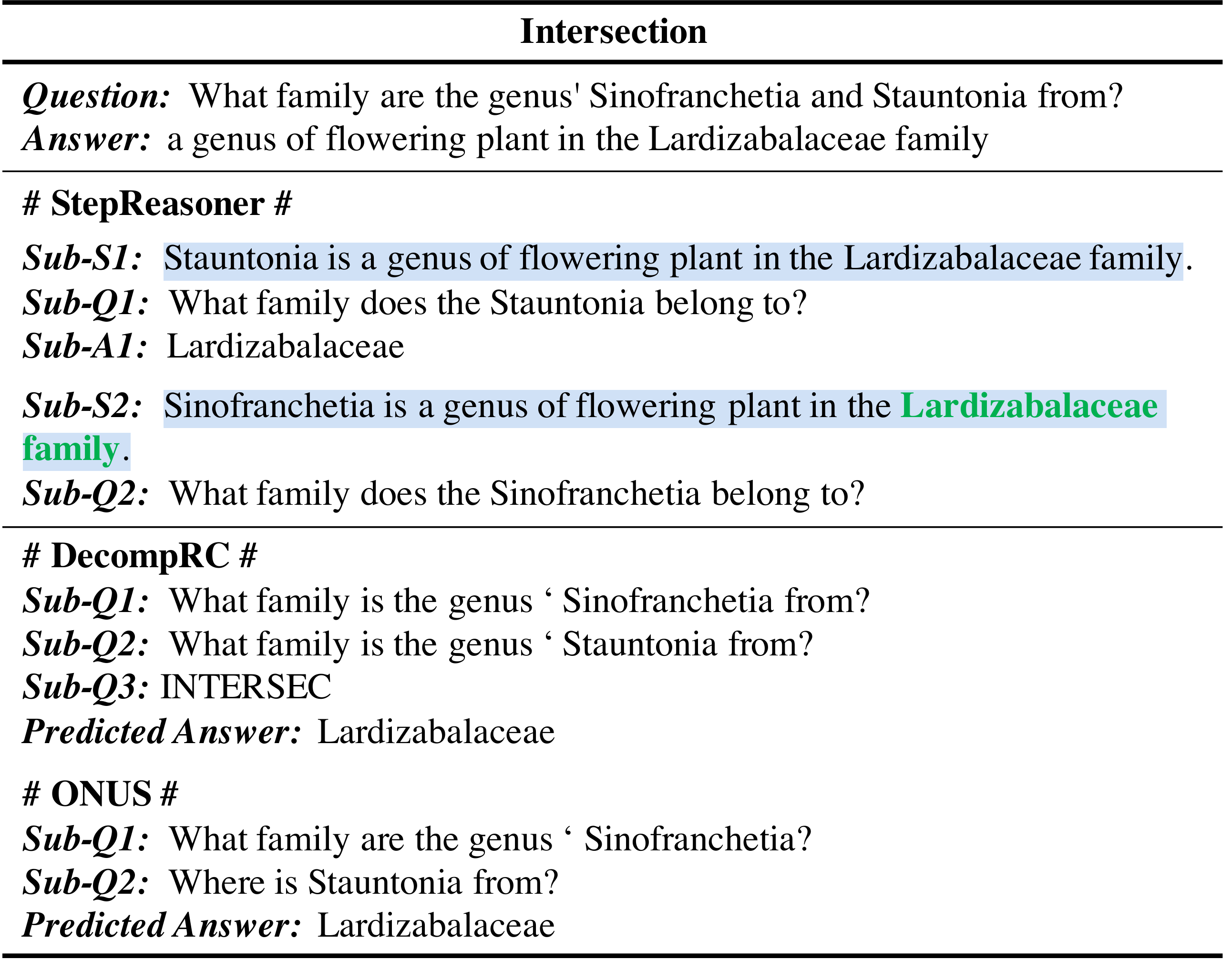} 
    \subcaption{An example of ``Intersection'' reasoning type.} 
    \label{figure1c}
  \end{minipage}
  
\caption{Three cases of other reasoning types. The \textcolor{newgreen}{green} phrases denote our predicted answers and the texts in shadow support the corresponding single-hop question generation. }
\label{figure_cases}
\end{figure}

We further present three cases of other reasoning types in Figure~\ref{figure_cases}, including ``Implicit-Bridge'', ``Comparison'' and ``Intersection''. We can see that the \emph{StepReasoner} generates high-quality decompositions for better interpretable multi-hop reasoning and predict accurate answers for all types compared to previous question decomposition based methods. 

For the ``Implicit-Bridge'' question in Figure~\ref{figure1a}, by first predicting the supporting sentences related to ``Sivarama Swami'' at the first hop, we can generate a sub-question to identify the implicit bridge ``Bhaktivedanta Manor'' for location query in the second hop. Although our predicted answer is different from the ground truth, it is also a reasonable response and more close to the golden one compared to the predictions of \emph{DecompRC} and \emph{ONUS}. These two methods both fail to decompose the multi-hop question and can only predict an intermediate answer.

For the ``Comparison'' and ``Intersection'' questions in Figure~\ref{figure1b} and~\ref{figure1c}, all methods predict the correct answers. However, we can generate more diverse single-hop sub-questions without requesting for any supervision, such as ``die'' and ``death'' for representing ``pass away'', and ``belong to'' for ``from''.
By contrast, the decompositions by \emph{DecompRC} are usually inflexibly from the original questions and the unsupervised \emph{ONUS} creates improper sub-questions with noises. We hope that combining our generated single-hop questions can also help to construct more natural and diverse multi-hop questions and further promote multi-hop reasoning performance.

\end{document}